\documentclass{article}

\usepackage{times}
\usepackage{graphicx} 
\usepackage{subfigure} 

\usepackage{natbib}
\usepackage{algorithm}
\usepackage{algorithmic}

\usepackage{hyperref}

\usepackage[accepted]{whi2016}

\icmltitlerunning{Meaningful Models}

\begin{document} 

\twocolumn[
\icmltitle{Meaningful Models: Utilizing Conceptual Structure to Improve Machine Learning Interpretability}

\icmlauthor{Nick Condry}{nscondry@gwmail.gwu.edu}
\icmladdress{The George Washington University,
            2121 I St. NW, Washington, DC 20052 USA}

\icmlkeywords{hmi, human-machine interaction, psychology, concept learning, machine learning, ICML}

\vskip 0.3in
]

\begin{abstract} 
The last decade has seen huge progress in the development of advanced machine learning models; however, those models are powerless unless human users can interpret them. Here we show how the mind`s construction of concepts and meaning can be used to create more interpretable machine learning models. By proposing a novel method of classifying concepts, in terms of `form' and `function', we elucidate the nature of meaning and offer proposals to improve model understandability. As machine learning begins to permeate daily life, interpretable models may serve as a bridge between domain-expert authors and non-expert users.
\end{abstract}

\section{Introduction}
In the last decade, machine learning algorithms have made huge strides, producing state-of-the-art results across a number of domains including image recognition, speech recognition, and natural language processing. However, while such results are exciting, there currently exists a gap between data modeling and knowledge extraction \cite{vellido2012making}. Machine learning models are rendered powerless unless they can be interpreted, thus in order for knowledge to be extracted from a model, we must account for the human cognitive factors involved in such a process. Interpretation must therefore be accounted for in machine learning processes, as shown in Figure 1. In addition to promoting more transparent results, interpretable models enable non-experts to utilize machine learning tools. For example, a business manager is more likely to accept a model`s recommendations if its results can be presented in business terms \cite{bose2001business}. As an ever-growing number of professionals come to rely on machine learning tools, the most successful models will provide an elegant user experience, presenting users with information and intelligence that are easily interpretable. 

\begin{figure}[ht]
\setlength{\belowcaptionskip}{-20pt}
\centering
\centerline{\includegraphics[width=\columnwidth]{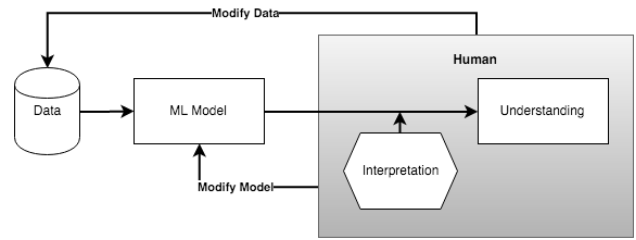}}
\caption{This illustration demonstrates the role of human interpretability in the development of a machine learning model. Without interpretable results, a human expert will not be able to accurately or efficiently modify their model or their datasets. This illustration is based on a diagram presented in \cite{vellido2012making}.}
\label{interpretation-process}
\end{figure} 

In the formal logic sense, an interpretation is a mapping of a formal construct to the entities and their relations it represents \cite{ruping2006learning}. Less formally, interpretability can be seen as a signaling problem; a model must present its output such that a specific meaning is conveyed to its user. To understand how to convey meaning, we must first understand the nature of meaning itself. Therefore, in order to design models for interpretability, we must first investigate the processes by which humans assign meaning to symbols, and how the mind extracts knowledge from information.

Whereas previous investigations into machine learning interpretability have largely focused on the relation between accuracy and interpretability, algorithm and feature selection, and model visualizations (e.g. \cite{ruping2006learning, ishibuchi2007analysis}, we will instead focus on the psychology of human concept learning. Using a relational model of meaning, we will propose a novel method of classifying concepts according to their structure and function within a given context. Based upon that method, we will offer several proposals to improve non-expert understanding of machine learning tools at a conceptual level. 

\section{Implicit Learning and Feature Extraction}
Humans are organisms that have evolved to learn from experience, evaluating novel stimuli through a process of comparison to previously stored stimuli. While learning, in the traditional sense of schooling and education, is an active process, in order to investigate the basis of knowledge, we`ll have to begin at the sub-conceptual and subconscious level.

The mind constantly and implicitly processes complex information in an incidental manner, without direct awareness of what has been learned \cite{seger1994implicit}. This process of passive knowledge acquisition is known as implicit learning.

Implicit learning began as a field of study with A.S. Reber`s work in the late 1960`s, and has been proposed as an evolutionary ancestor of explicit thought \cite{reber1967implicit, reber1992cognitive}. This process occurs automatically, and represents the subtle yet constant re-wiring of a brain`s neurons as they adapt in response to new stimuli \cite{sanders1987frequency}. Most importantly, implicit learning occurs at the subconscious, or pre-conscious level; therefore, the knowledge gained is sub-conceptual, which is to say, the patterns learned are not immediately associated with a reference symbol \cite{kihlstrom1987cognitive}. Instead, this process extracts relevant features from the local environment via the mind`s lower level perceptual processes \cite{schyns1998development}. A feature is an individual measurable property of a phenomenon being observed \cite{bishop2006pattern}. Features may be continuous or categorical, and they comprise the most basic building block of human knowledge \cite{schyns1998development}.

\section{From Features to Concepts}
The process of feature extraction is constant and unconscious; to bring this knowledge into the conscious domain requires conceptualization \cite{goodman2008rational}. A concept is an abstract system composed of a set of features paired with a symbolic representation. In many ways, conceptualization mirrors a simple dictionary structure, where the symbol acts as the key, and its associated feature set is the value. The symbolic representation can be any real or abstract token, including images, sounds, and smells. However, the most common form of symbolic representation is a word, a character or combination of characters. For example, the concept of a dog might contain the features [furry: yes, ears: 2, legs: 4, tail: yes] and would be denoted by the character string: `dog`. Since concepts are composed of a multi-dimensional set of features, they are inherently complex symbolic objects. 

Concepts are abstract, meaning they can be applied to novel stimuli, and concept learning relies on incremental assumptions \cite{katz2007issues}. The mind, as a concept formation system, accepts a stream of observations (i.e. events, objects, instances), and discovers a classification scheme over the data stream. Learning occurs not as a single event but as a continuous process; the mind`s classification scheme evolves and changes as new observations are processed \cite{Fisher14}. Figure 2 \cite{Dietterich82} demonstrates this incremental learning process by which an agent adapts to its environment, organizing experiences to improve its performance \cite{Fisher14}.

\begin{figure}[ht]
\vskip 0.2in
\begin{center}
\centerline{\includegraphics[width=\columnwidth]{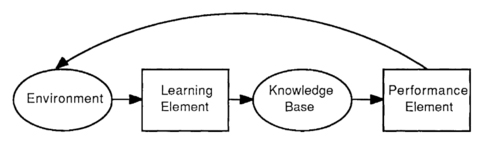}}
\caption{This flow chart illustrates the act of learning as a continuous incremental process, by which an organism adapts to improve its fitness within a given environment.}
\label{learning_process}
\end{center}
\vskip -0.2in
\end{figure}

This view of learning demonstrates that learning is not a discrete act, but rather a continuous process by which new information contributes to the evolution of existing concepts and the formation of new concepts. Furthermore, it aligns with \cite{ruping2006learning} heuristic of interpretability, which states, “people tend to find those things understandable, that they already know”. Thus, when building a meaningful model, the intended audience must be taken into account when structuring output. If the output can be phrased or structured in a familiar way, subjects will be more likely to implicitly trust and utilize the information.

\section{From Concepts to Meaning}
Having established concepts as a system composed of a [key, value] pair, where the key is a symbol and the value is the associated feature set, we can look at meaning. The word “meaning” is often used in a variety of ways, from Plato`s physically irreducible mystical essences to ideas of how words are used \cite{ludwig1953philosophical}. Here, I will offer a view which finds its roots in connectionist psychological models, but until recently was unrealized at scale \cite{o2000computational}. This view holds that since words simply denote clusters of features, words themselves have no inherent meaning; stripped of its associated features, a word is simply a meaningless symbol. Instead, meaning arises from the cognitive mapping of a word (or symbol) onto an underlying feature map \cite{landauer2013handbook}. For example, to someone with no knowledge of the English language, the word `tree' would mean nothing, as their mind has not mapped the symbol to a set of features. However, to a native speaker, not only would `tree' have meaning, but they could likely identify `forest' as a similar concept, due to their overlapping feature sets. This theory of meaning has gained validation from the rise of latent semantic analysis (LSA) techniques, which construct models from the implicit relational mapping of a text. This `map` does not exist in itself, it is an abstraction – an infinite number of point-to-point distances computed by triangulation from earlier established points \cite{landauer2013handbook}. However, models created in the manner have proven highly accurate, and overlaying word-symbols on top of such maps have produced highly intuitive results.

Using this approach, we can view `meaning` as a fundamentally relational property, as a word`s relation to the semantic system in which it exists defines its meaning. Importantly, this leads us to realize that to efficiently convey meaning, we must start at the sub-conceptual level by identifying the specific information we hope to convey, then crafting a message such that it conveys the intended features given the context of audience. Given this theory, I will use the word ``meaning'' as shorthand for ``the set of features associated with a symbol, given context''.

\section{The Form and Function of Concepts}
The relational theory of meaning holds that a symbol, say, a word or an image, may hold different meanings in different contexts, given that it interacts with those contexts differently. While this might seem to imply that words cannot be assigned any true meaning, in practice this is not the case. Through shared communication protocols such as language, individually relative meanings solidify into a statistically canonical cultural form \cite{goldstone2002using}.

Nevertheless, this theory lacks a direct explanation of the relationship between a symbol and meaning. We posit that this relationship can be best understood in terms of \textit{form} and \textit{function}. The \textit{function} of a concept is its meaning, given context, and it represents how the concept interacts with its larger semantic context. Concepts that share their \textit{function} are synonyms \cite{kao2007natural}. The \textit{form} is the specific instance of the class of objects defined by the object's function. For example, compare the following three phrases, ``I`m going to the store'', ``I`m heading to the store'', and ``I`m heading the soccer ball''. Given the context of the first two phrases, ``going'' and ``heading'' share the same meaning, and can thus be considered different \textit{forms}, or instances, of the same conceptual \textit{function}, or class. Given the context of the second two phrases, the conceptual \textit{form}, ``heading'', is the same, but its \textit{function} differs.

In some aspects, this categorization of concepts by form and function represents an extension of the ``theory theory'' of concepts in which concepts are composed of \textit{core} and \textit{peripheral} features (see: \cite{carey1985conceptual, gopnik1997words}. An object's \textit{core features} are its causally deepest properties, whereas \textit{peripheral features} refer to incidental features of a concept that do not directly define its nature. These descriptions of features as either core or peripheral are useful in qualitative description, but are difficult to translate into more technical contexts. Instead, we propose that \textit{function} best encapsulates the meaning of \textit{core features}, and \textit{form} best encapsulates the meaning of \textit{peripheral features}. The \textit{essence}, or core of a concept, is its meaning, defined by the concept's function within a given context. Peripheral qualities, or form, are in turn best understood as the characteristics of a specific object. For example, within the simple context presented in Figure 3, the rock interacts with a piece of paper by resting on top of it. While the \textit{form} of the rock may be a small, grey, 2lb stone, within the given context, its \textit{function} is to apply downward force on the paper, therefore its \textit{meaning} is `paper weight'. Similarly, as the paper supports the rock, from the rock's perspective, the \textit{function} of the paper is support, so its \textit{meaning} is ground. \textit{Forms} can change without altering the operation of a system, so long as the object retains it's \textit{function}.

Since an object's meaning is defined by interaction with its context, and the interaction can be viewed as a function, a relationship between inputs and output, meaning can be understood as a function within a larger process of interaction.

\begin{figure}[ht]
\vskip 0.2in
\begin{center}
\centerline{\includegraphics[width=\columnwidth]{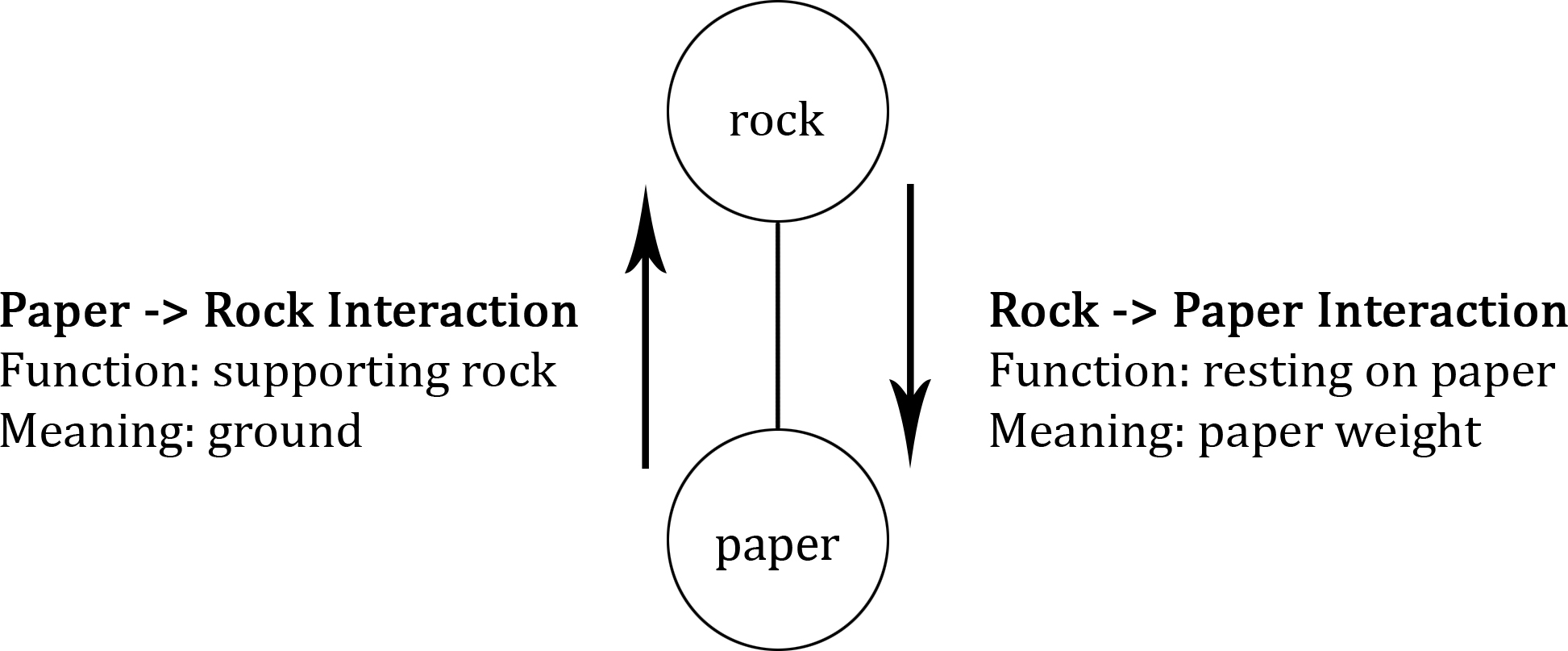}}
\caption{This diagram displays how meaning arises through interaction. This diagram also reveals that meaning is a function of perspective: from the perspective of the paper, the rock is simply a weight whereas to the rock, the paper may as well be the ground.}
\label{paper_rock}
\setlength{\belowcaptionskip}{-20pt}
\end{center}
\vskip -0.2in
\end{figure} 

\section{Proposals to Improve Meaningful Models}
\subsection{Clearly Outline a Model's \emph{Function}}
The function of a concept is defined by the change it enacts on its context, and thus represents a transition from an initial state to an output state. Thus, to improve model interpretability, models should have very clearly defined requirements for input and the goals of the output. For example, doctors might be supplied with a few models that perform different tasks, including quality-of-life (QOL) assessments, anomaly detection, and DNA sequence mining \cite{cleophas2013machine}. Authors of such models should clearly state the purpose and intended applications of their work. If the model is only intended to perform exploratory data analysis, the author should emphasize in their discussion that confirmatory data analysis is required \cite{foster2014machine}. Furthermore, authors should directly address the transportability of the model, i.e. which aspects of the method can be directly used in novel situations, and which aspects must be tuned for further application.

Additionally, authors should minimize the number of attributes in their classifiers. Minimizing attributes creates a simpler, and therefore more interpretable, form of the model, and also decreases the risk of overfitting, especially in smaller studies. One approach to limit attributes might involve variable ranking (see: \cite{bekkerman2003distributional, caruana2003benefitting, forman2003extensive, weston2003use}). Another viable method proposed by \cite{weigend1990generalization} pares down variables using a weight elimination algorithm.

\subsection{Place the Model in Context}
In addition to specifying the purpose and scope-of-use of a model, authors should attempt to construct models such that they complement and expedite existing processes. In doing so, the “meaning” of their model will be elucidated by its context in the existing process. For example, in the early stages of developing a medical diagnostic imaging application, it is impossible to conclusively prove that the application works, but possible to prove that it does not work \cite{foster2014machine}. If the latter is the case, it is best to discover such quickly, so that new processes and applications may be developed. A model in this process would become more meaningful, by virtue of having a clearly defined function within the scope of a larger system. Additionally, incorporating models into existing processes forces those models to incorporate some level of domain knowledge, and serve as useful tools rather than complete solutions unto themselves.

\subsection{Design for User Experience}
Finally, when developing models that aim to solve specific problems within a given domain area, thought should be given to preparing a front-end for users within that domain. A well-designed front-end would ideally accomplish the above proposals by clearly specifying required inputs, presenting coherent outputs, and positioning the model as a tool within a larger process or framework. Current developments of machine learning platforms such as Google Cloud Platform, Amazon Machine Learning, Microsoft Azure, and H20.ai have made strong progress in this regard, combining powerful models with intuitive representations. 

While the algorithms and structure of the model itself accounts for the model`s function, a cohesive front-end provides an overlaid form for the information conveyed. Essentially, this front-end can be viewed as a translation between the direct model output and a non-expert user. This translation should capitalize on the fundamentals of human concept acquisition by providing both information in a familiar format, and context. To this end, authors should focus on key user experience metrics, such as: will the users recommend the tool? Does this tool create a more efficient or effective process? What are the most significant usability problems with the tool? Are usability improvements being made from one version to the next \cite{albert2013measuring}? These questions place an emphasis on considering the understandability of a model in the design of the algorithms. Interpretability is difficult to achieve as a post-processing step; the relationship between understandability and accuracy must be accounted for from the start \cite{ruping2006learning}.

\section{Conclusion}
We have analyzed the psychology of human concept learning, and identified how the mind`s construction of concepts and meaning can be used to create more interpretable machine learning models. Meaning arises from the interaction of a concept within a specified context. Furthermore, the identity of an object and its meaning can be fully described by two traits: form and function. Form describes the exact qualities and structure of an object, while function describes the object`s meaning as a function of its interaction in its context. Furthermore, this promotes a view of concepts as functions in context, which allows them to be conceptualized as a relationship between input and output.

Thus, the interpretability of a model on a conceptual level can be bolstered by clearly indicating the model`s input requirements and output goals, and providing context for the model within a larger process. Additionally, these goals may be combined through the development of a cohesive front-end to present information in a familiar format and expand the usability of a model to non-expert users. 


\bibliography{meaning_trim.bib}

\end{document}